\documentclass[sigconf,natbib=true]{acmart}
\usepackage[ruled,vlined]{algorithm2e}
\usepackage{amsmath}
\usepackage{multirow}
\usepackage{mathtools}
\usepackage{array}
\usepackage{geometry}
\usepackage{subfig}
\usepackage{xspace}
\usepackage{algorithmic}
\usepackage{subfig} 
\usepackage{float} 
\usepackage[font=bf]{caption}
\newcommand{\spara}[1]{\vspace{1mm}\noindent\textbf{#1.}}
\newcommand{\DART}{\textsf{DART}\xspace}

\AtBeginDocument{%
  }

\setcopyright{acmlicensed}
\copyrightyear{2026}
\acmYear{2026}
\acmDOI{XXXXXXX.XXXXXXX}

\acmConference[SIGIR'26]{International ACM SIGIR Conference}{July 20--24}{Melbourne | Naarm, Australia}

\acmISBN{978-1-4503-XXXX-X/18/06}

\begin{document}

%%
%% The "title" command has an optional parameter,
%% allowing the author to define a "short title" to be used in page headers.
\title{Mitigating Structural Overfitting: A Distribution-Aware Rectification Framework for Missing Feature Imputation}

\author{Yifan Song}
\authornote{These authors contributed equally to this research.}
\affiliation{%
  \institution{HKUST(GZ)}
  \country{}
}
\email{ysong853@connect.hkust-gz.edu.cn}

\author{Fenglin Yu}
\authornotemark[1]
\affiliation{%
  \institution{Carnegie Mellon University}
  \country{}
}
\email{mikukuovo@gmail.com}

\author{Yihong Luo}
\authornotemark[1]
\affiliation{%
  \institution{HKUST}
  \country{}
}
\email{yluocg@connect.ust.hk}

\author{Xingjian Tao}
\affiliation{%
  \institution{HKUST(GZ)}
  \country{}
}
\email{mikukuovo@gmail.com}

\author{Siya Qiu}
\affiliation{%
  \institution{HKUST}
  \country{}
}
\email{sqiual@connect.ust.hk}

\author{Kai Han}
\affiliation{%
  \institution{Shanghai University of Finance and Economics}
  \country{}
}
\email{hankai@mail.shufe.edu.cn}

\author{Jing Tang}
\affiliation{%
  \institution{HKUST(GZ) \& HKUST}
  \country{}
}
\email{jingtang@ust.hk}

\renewcommand{\shortauthors}{Yifan Song et al.}

\begin{abstract}

Incomplete node features are ubiquitous in real-world scenarios such as user profiling and cold-start recommendation, which severely hinders the practical deployment of graph learning systems (e.g., GNNs). Existing solutions typically rely on diffusion-based structural smoothing (e.g., feature propagation) to impute missing values. However, we find that these approaches suffer from structural overfitting, leading to three progressive challenges: 1) performance degradation on disjoint graphs, 2) loss of semantic diversity due to over-smoothing, and 3) feature distribution shift when generalizing to unseen graph structures (inductive tasks). To address these challenges, we introduce the \textbf{\DART} framework. It begins by employing {\em Global Structural Augmentation (GSA)}, which establishes global correlations to bridge disjoint components and extend diffusion coverage. Building upon this, we design a semantic rectifier based on masked autoencoding. This module learns the latent feature manifold to recover natural semantic details. Crucially, we introduce a test-time distribution rectification mechanism that projects structurally biased features back onto the learned manifold during inference, effectively bridging the inductive distribution gap. Furthermore, considering that synthetic masking fails to reflect real-world sparsity, we present a new dataset \textbf{Sailing} collected from voyage records with naturally missing attributes. Extensive experiments on six public datasets and Sailing demonstrate that \DART significantly outperforms state-of-the-art methods in both transductive and inductive settings. Our code and dataset are available at \url{https://github.com/yfsong00/DART}.
\end{abstract}

%%
%% The code below is generated by the tool at http://dl.acm.org/ccs.cfm.
%% Please copy and paste the code instead of the example below.
%%
% \begin{CCSXML}
% <ccs2012>
%    <concept>
%        <concept_id>10002951.10003227.10003351</concept_id>
%        <concept_desc>Information systems~Data mining</concept_desc>
%        <concept_significance>500</concept_significance>
%        </concept>
%    <concept>
%        <concept_id>10010147.10010257.10010321</concept_id>
%        <concept_desc>Computing methodologies~Machine learning algorithms</concept_desc>
%        <concept_significance>500</concept_significance>
%        </concept>
%  </ccs2012>
% \end{CCSXML}

% \ccsdesc[500]{Information systems~Data mining}
% \ccsdesc[500]{Computing methodologies~Machine learning algorithms}

%%
%% Keywords. The author(s) should pick words that accurately describe
%% the work being presented. Separate the keywords with commas.
%% A "teaser" image appears between the author and affiliation
%% information and the body of the document, and typically spans the
%% page.
% \begin{teaserfigure}
%   \includegraphics[width=\textwidth]{sampleteaser}
%   \caption{Seattle Mariners at Spring Training, 2010.}
%   \Description{Enjoying the baseball game from the third-base
%   seats. Ichiro Suzuki preparing to bat.}
%   \label{fig:teaser}
% \end{teaserfigure}

% \received{20 February 2007}
% \received[revised]{12 March 2009}
% \received[accepted]{5 June 2009}

%%
%% This command processes the author and affiliation and title
%% information and builds the first part of the formatted document.
\maketitle

\begin{sloppy}
\section{Introduction}
\label{sec:introduction}

Graphs are important structures in the real world, which are capable of modeling complex interactions among various objects. Graph learning approaches such as graph neural networks (\textbf{GNNs})~\cite{kipf2016semi,cai2018comprehensive,fang2023spatio,huang2022auc,jiang2019semi,henaff2015deep,velickovic2017graph,lsgnn,degem,fgsam} have demonstrated impressive effectiveness in a wide range of applications for graph analysis, including trajectory prediction~\cite{mohamed2020social,li2019grip} and user classification~\cite{wang2021mixup,kipf2016semi,xiao2022graph,wang2020nodeaug}. These methods typically achieve robust embeddings by combining graph feature information and graph structure. However, a critical bottleneck in real-world deployments is the ubiquity of \emph{incomplete node features}. A prominent example arises in the domain of maritime navigation, where critical features such as vessel velocity and wind conditions may be incomplete or missing due to signal interference or sensor limitations, seriously degrading the performance of graph learning methods~\cite{rossi2022unreasonable,um2023confidencebased}.

To address this issue, recent studies have proposed various imputation techniques~\cite{huo2023t2,taguchi2021graph,jiang2020incomplete}. Among them, diffusion-based methods based on feature propagation (FP)~\cite{rossi2022unreasonable} have achieved superior performance by iteratively diffusing known features over the graph topology to reconstruct missing features. However, we argue that they fundamentally suffer from \emph{structural overfitting}, where imputed values are deterministically conditioned on graph topology while ignoring the intrinsic feature manifold. This dependency introduces three critical challenges: 1) The imputation quality is highly sensitive to graph connectivity. As illustrated in Figure \ref{fig:connected}, previous studies only validate their effectiveness on the largest connected component, which is unreasonable. The performance of diffusion-based methods (FP and PCFI) is significantly degraded on whole graphs with multiple connected components. 2) These methods face over-smoothing problem in transductive tasks. Figure \ref{fig:diversity} reports the MAD metric \cite{chen2020measuring}, the small values of diffusion-based methods indicate a severe reduction in feature diversity. 3) Structural overfitting leads a feature distribution shift between training and inference in inductive tasks. As shown in Figure \ref{fig:FP_problem}, the structure of the partial training graph differs from the complete inference graph. It causes the imputed feature distributions to diverge, thereby affecting the generalization performance in downstream tasks.

\begin{figure}[t]
\centering
\captionsetup[subfloat]{captionskip=0.5mm}
\begin{small}
\begin{tabular}{ccc}
\hspace{-2mm}\subfloat[{Performance decline of diffusion-based methods on Cora.}]{\includegraphics[width=0.48\linewidth]{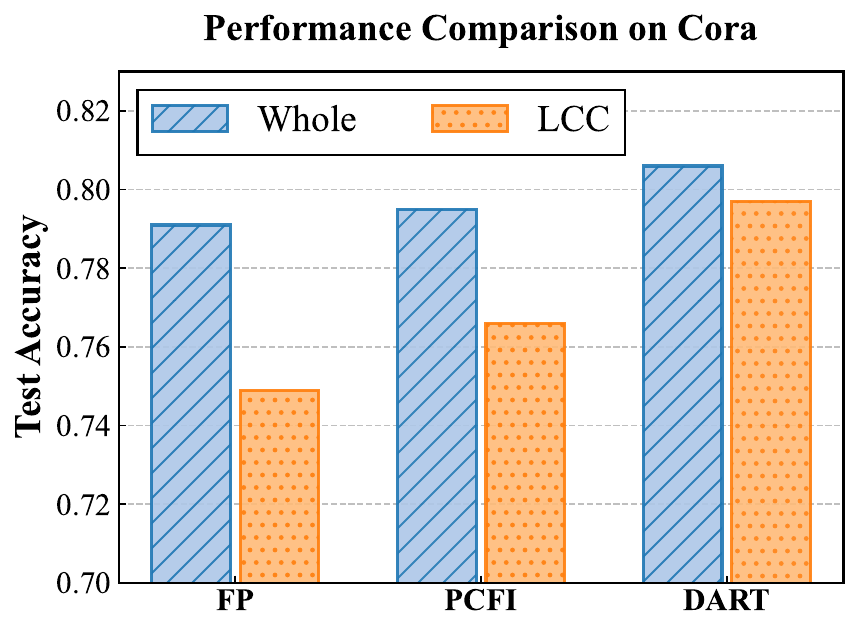}\label{fig:connected}} &
\hspace{-2mm}\subfloat[{MAD metric (small value means over-smoothing).}]{\includegraphics[width=0.48\linewidth]{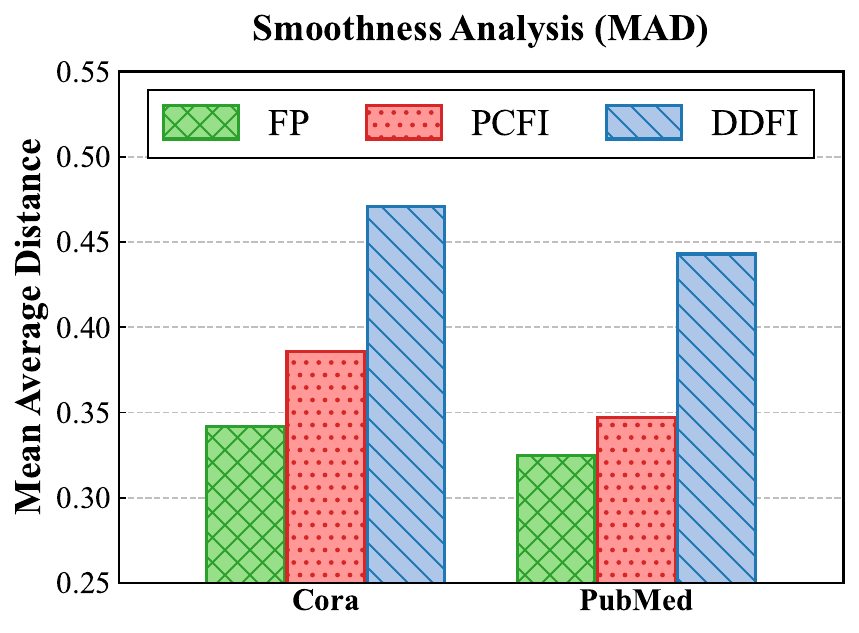}\label{fig:diversity}} &
\end{tabular}
\vspace{-2mm}
\caption{Analysis of Structural Overfitting in FP-based methods. LCC means validate on the largest connected component (previous work's unreasonable setting) while Whole means validate on whole graph (our setting).}
\label{fig:problem}
\end{small}
\vspace{-4mm}
\end{figure}

To tackle these challenges, we propose \textbf{\DART} (\textbf{D}istribution-\textbf{A}ware \textbf{R}ectification via \textbf{T}est-time adaptation), a framework that decouples structural estimation from semantic manifold learning. First, we design \textbf{Global Structural Augmentation (GSA)} that utilizes label consistency to establish semantic bridges between disjoint components, which extends the diffusion scope. Second, to restore semantic diversity, we train a masked autoencoder as the manifold learner to reconstruct features from structurally perturbed inputs. Crucially, a Gaussian-based masking strategy is used to force the model to capture the robust feature manifold rather than learning smoothed values. Finally, to bridge the inductive gap, we introduce a test-time distribution rectification mechanism. During inference, we project the structurally biased backs onto the learned manifold via the encoder-decoder framework, effectively correcting the distribution shift.

Furthermore, existing methods typically evaluate performance by randomly masking features in complete datasets (synthetic missing), which fails to reflect the complexity of real-world sparsity. To benchmark imputation methods under realistic scenes, we present a new dataset called \textbf{Sailing} collected from real maritime voyage records provided by the Danish Maritime Authority (DMA)~\cite{dma}. Notably, 80.4\% of the features in Sailing are \emph{naturally missing}, providing a rigorous dataset for real-world deployment.

In summary, our main contributions are:
\begin{itemize}
    \item We identify the structural overfitting issue in existing diffusion-based methods, which leads to limited performance on disjoint graphs, low diversity and distribution shift.
    \item We propose \DART, a novel framework that integrates a manifold learner with test-time distribution rectification, effectively improving feature diversity and mitigating inductive distribution shift.
    \item We release Sailing, a real-world maritime dataset with naturally missing attributes. Extensive experiments on Sailing and six public datasets demonstrate that \DART significantly outperforms state-of-the-art methods.
\end{itemize}

\begin{figure}[t] 
\centering 
\includegraphics[width=\linewidth]{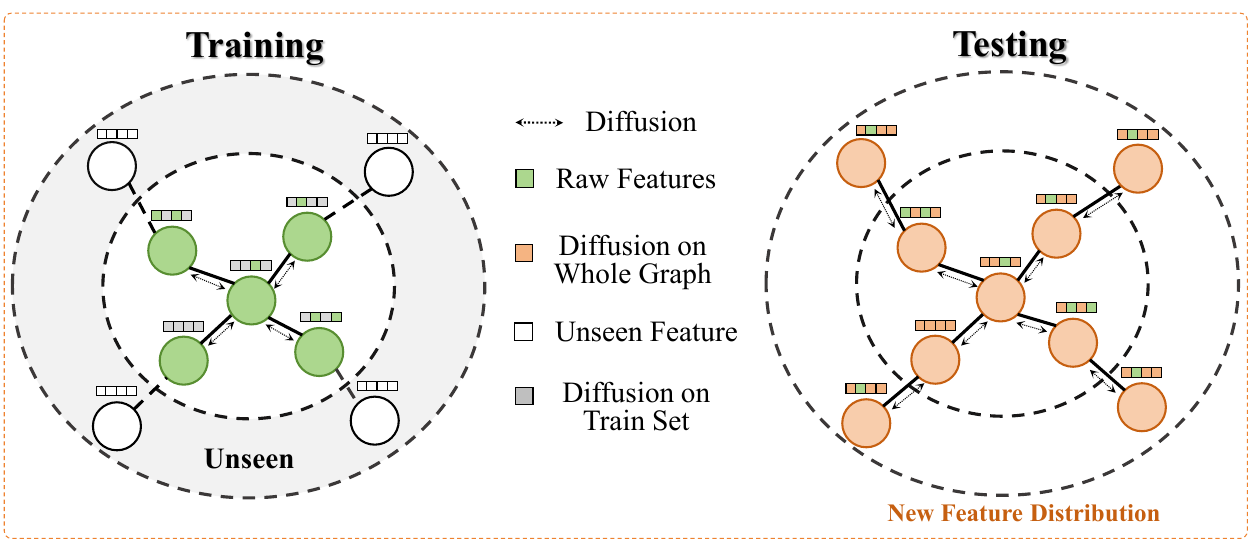}  
\vspace{-3mm}
\caption{Illustration of Inductive Distribution Shift caused by structural changes.} \label{fig:FP_problem}
\vspace{-4mm}
\end{figure}

\section{Preliminaries}
\label{sec:preliminaries}
\subsection{Notations and Problem Formulation}
Let $G=(\mathcal{V}, \mathcal{E})$ be an undirected graph with $|\mathcal{V}|=N$ nodes. We denote the adjacency matrix as $\mathbf{A} \in \mathbb{R}^{N\times N}$ and the node label vector as $\mathbf{Y} \in \mathbb{R}^{N}$. The node feature matrix is denoted as $\mathbf{X} \in \mathbb{R}^{N\times d}$. Consider a realistic scenario where node features are partially observed. Let $\mathbf{M} \in \{0, 1\}^{N \times d}$ be a binary mask matrix, where $\mathbf{M}_{ij}=1$ indicates the $j$-th feature of node $v_i$ is observed, and 0 otherwise. The observed feature matrix is given by $\mathbf{X}_{obs} = \mathbf{X} \odot \mathbf{M}$.

\spara{Problem Formulation} The goal of missing feature imputation is to learn a mapping function $\Phi: (\mathbf{X}_{obs}, \mathbf{A}) \to \hat{\mathbf{X}}$ that reconstructs the complete feature matrix $\hat{\mathbf{X}}$. The imputed features are then used by a downstream model $f_\theta$ (e.g., a GNN) to minimize the task-specific loss (e.g., cross-entropy for node classification):
\begin{equation}
    \min_{\Phi, \theta} \mathcal{L}_{task}(f_\theta(\Phi(\mathbf{X}_{obs}, \mathbf{A}), \mathbf{A}), \mathbf{Y}).
\end{equation}
We consider both \textit{transductive} settings (where $\mathbf{A}_{test} = \mathbf{A}_{train}$) and \textit{inductive} settings (where $\mathbf{A}_{test}$ contains unseen nodes and edges disjoint from $\mathbf{A}_{train}$).

\spara{Inductive Feature Distribution Shift} 
In inductive graph learning, the structural contexts differ significantly between training and testing. We denote the \textit{feature distribution shift} $\Delta_{\mathcal{S}}$ in inductive tasks as the discrepancy between the distributions of features imputed strictly via structural propagation on disjoint topologies. Formally, let $P_{\Phi}(\hat{\mathbf{X}}|\mathbf{A})$ denote the conditional distribution of features reconstructed by a structure-dependent mapping $\Phi$ (e.g., FP). The shift is defined as:
\begin{equation}
\label{eq:delta}
    \Delta_{\mathcal{S}} = \mathcal{F}\left( P_{\Phi}(\hat{\mathbf{X}} | \mathbf{A}_{test}) \parallel P_{\Phi}(\hat{\mathbf{X}} | \mathbf{A}_{train}) \right),
\end{equation}
where $\mathcal{F}$ is a divergence measure (e.g., Kullback-Leibler divergence). Crucially, since $P_{\Phi}$ is conditioned on the graph topology, structural differences between $\mathbf{A}_{train}$ and $\mathbf{A}_{test}$ introduce a structural bias $\delta_{struct}$, causing the imputed support to deviate from the true manifold $\mathcal{M}$ (i.e., introducing "off-manifold" noise), which serves as the primary challenge we aim to rectify.

\subsection{Revisiting Feature Propagation}
\label{sec:revisiting_fp}

Feature Propagation (FP)~\cite{rossi2022unreasonable} serves as the foundational diffusion mechanism for imputing the missing node features. It assumes that valid feature signals should vary smoothly across the topological structure. Here we formulate FP as a discrete Dirichlet boundary value problem on the graph. Specifically, the reconstruction of missing features aims to minimize the graph Dirichlet energy while strictly adhering to the observed signals:
\begin{equation}
    \label{eq:fp_energy}
    \mathcal{E}(\hat{\mathbf{X}}) = \text{tr}(\hat{\mathbf{X}}^\top \tilde{\mathbf{L}} \hat{\mathbf{X}}), \quad \text{s.t. } \hat{\mathbf{X}}_{\mathcal{V}_{obs}} = \mathbf{X}_{obs},
\end{equation}
where $\tilde{\mathbf{L}} = \mathbf{I} - \tilde{\mathbf{A}}$ denotes the normalized graph Laplacian, and the constraint $\hat{\mathbf{X}}_{\mathcal{V}_{obs}} = \mathbf{X}_{obs}$ ensures fidelity to the ground truth. The optimal solution corresponds to the steady state of a heat diffusion process constrained by boundary conditions. In practice, this harmonic solution is efficiently approximated via an iterative diffusion-and-reset procedure:
\begin{equation}
    \label{eq:fp_update}
    \mathbf{X}^{(k+1)} = (\mathbf{1} - \mathbf{M}) \odot (\tilde{\mathbf{A}} \mathbf{X}^{(k)}) + \mathbf{M} \odot \mathbf{X}_{init},
\end{equation}
where $\mathbf{X}^{(0)} = \mathbf{X}_{init}$ (typically zero-filled). In each iteration, the feature information propagates from neighbors through the normalized adjacency $\tilde{\mathbf{A}}$, effectively acting as a low-pass filter that smooths high-frequency signals. Meanwhile, the term $\mathbf{M} \odot \mathbf{X}_{init}$ enforces the boundary condition by explicitly resetting the observed nodes to their original values. This process repeats until convergence, yielding the structurally smoothed estimation $\hat{\mathbf{X}}_{fp}$. These issues indicate that FP-based imputations are inherently biased towards the current graph topology and ignore the latent manifold of the real features, which motivates our proposed \DART framework.
\section{The \DART Framework}
\label{sec:methodology}

To tackle the above challenges of previous feature imputation methods, we propose \DART, a framework combines global structural augmentation with a masked autoencoder to recover semantic manifolds and addresses the distribution shift in graph feature imputation through a novel test-time adaptation strategy. Unlike previous methods that strictly rely on structural diffusion, \DART functions as a decoupling framework that separates \textit{structural estimation} from \textit{semantic manifold learning}. As illustrated in Figure~\ref{fig:FPMAE}, the overall workflow consists of three progressive stages.

First, \textbf{Global Structural Augmentation (GSA)} constructs a global affinity topology to bridge disconnected components, ensuring valid structural initialization. Second, a \textbf{Self-Supervised Manifold Learner} utilizes a graph-based masked autoencoder with continuous semantic perturbation to capture the intrinsic feature distribution, thereby recovering semantic diversity. Finally, to bridge the generalization gap in inductive tasks, we introduce a \textbf{Test-Time Distribution Rectification} mechanism, which projects structurally biased estimates back onto the learned manifold during inference to align the feature distribution.

\begin{figure*}[h] 
\centering 
\includegraphics[width=\linewidth]{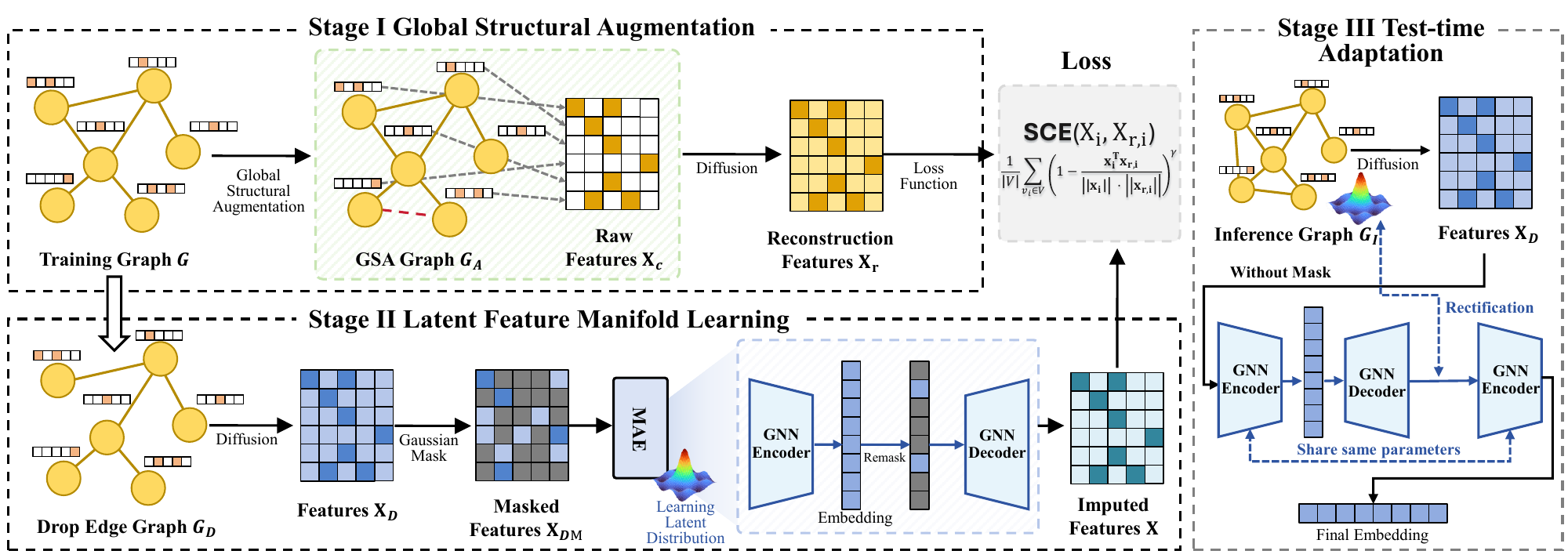}  
\vspace{-5mm}
\caption{Training and inference process of \DART.} \label{fig:FPMAE} 
\vspace{-3mm}
\end{figure*}

\subsection{Stage I: Global Structural Augmentation}
\label{sec:gsa}

\begin{algorithm}[t]
\caption{Global Structural Augmentation (GSA)}
\label{alg:gsa}
\KwIn{Training Graph $\mathcal{G}=(\mathcal{V}, \mathcal{E})$, Features $\mathbf{X}$, Labels $\mathbf{Y}_{train}$, Parameters $k, \tau, \delta$.}
\KwOut{Augmented Graph $\tilde{\mathcal{G}}=(\mathcal{V}, \tilde{\mathcal{E}})$.}
\BlankLine
$\tilde{\mathcal{E}} \gets \mathcal{E}$\;
\ForEach{node $v_i \in \mathcal{V}$}{
    \tcp{1. Similarity Computation}
    Compute weights: $w_j \gets \text{sim}(\mathbf{x}_i, \mathbf{x}_j), \forall v_j \in \mathcal{V}$\;
    Normalize probability: $p_j \gets \frac{\exp(w_j / \tau)}{\sum_{v_l \in \mathcal{V}} \exp(w_l / \tau)}$\;

    \tcp{2. Semantically-Guided Linking}
    Sample set $\mathcal{S}_i$ of size $k$ from $\mathcal{V}$ with distribution $\{p_j\}$\;
    \ForEach{node $v_j \in \mathcal{S}_i$}{
    \tcp{Filter by label consistency}
    $\mathcal{V}_{link} \gets \{ v_j \in \mathcal{S}_i \mid y_j = y_i \}$\;
    \tcp{Filter by similarity threshold}
    $\mathcal{V}_{link} \gets \{v_j \in \mathcal{S}_i \mid w_j > \delta \cap v_j \notin \mathcal{V}_{link} \}$\;
    $\tilde{\mathcal{E}} \gets \tilde{\mathcal{E}} \cup (\{v_i\} \times \mathcal{V}_{link})$\;
    }
}
\KwRet{$(\mathcal{V}, \tilde{\mathcal{E}})$}\;
\end{algorithm}

As analyzed in Section \ref{sec:preliminaries}, diffusion-based methods rely on valid propagation paths to impute missing features. We formalize the relationship between topological connectivity and imputation effectiveness with the following lemma:

\begin{lemma} 
\label{lemma:reachability}
(Diffusion Reachability Bound). Let $\mathbf{S} = \mathbf{D}^{-1/2}\mathbf{A}\mathbf{D}^{-1/2}$ be the normalized adjacency matrix. The imputation of a missing node $v_i$ via feature propagation is theoretically given by a weighted sum of observed features over all possible paths. If $v_i$ belongs to a connected component with no observed nodes (i.e., disjoint from $\mathcal{V}_{obs}$), the imputed value is strictly zero, leading to maximum error.
\end{lemma}

\begin{proof}
Denote $\mathcal{O}$ and $\mathcal{U}$ as the sets of observed and missing nodes respectively, we first partition the matrix $\mathbf{S}$ and feature vector $\mathbf{X}$ according to $\mathcal{O}$ and $\mathcal{U}$:
\begin{equation}
    \mathbf{S} = \begin{bmatrix} \mathbf{S}_{\mathcal{O}\mathcal{O}} & \mathbf{S}_{\mathcal{O}\mathcal{U}} \\ \mathbf{S}_{\mathcal{U}\mathcal{O}} & \mathbf{S}_{\mathcal{U}\mathcal{U}} \end{bmatrix}, \quad 
    \mathbf{X} = \begin{bmatrix} \mathbf{X}_{\mathcal{O}} \\ \mathbf{X}_{\mathcal{U}} \end{bmatrix}.
\end{equation}
The optimal solution of feature propagation for the missing nodes $\mathbf{X}_{\mathcal{U}}$ is obtained by setting the partial derivative of the Dirichlet energy with respect to $\mathbf{X}_{\mathcal{U}}$ to zero, yielding the closed-form harmonic solution:
\begin{equation}
    \mathbf{X}_{\mathcal{U}} = (\mathbf{I} - \mathbf{S}_{\mathcal{U}\mathcal{U}})^{-1} \mathbf{S}_{\mathcal{U}\mathcal{O}} \mathbf{X}_{\mathcal{O}}.
\end{equation}
Since the spectral radius $\rho(\mathbf{S}_{\mathcal{U}\mathcal{U}}) < 1$, we can expand the inverse term using the Neumann Series. For a specific missing node $v_i \in \mathcal{U}$, its imputed value is:
\begin{equation}
    \hat{\mathbf{x}}_i = \sum_{v_j \in \mathcal{O}} \left( \sum_{k=0}^{\infty} \sum_{v_m \in \mathcal{U}} (\mathbf{S}_{\mathcal{U}\mathcal{U}}^k)_{im} (\mathbf{S}_{\mathcal{U}\mathcal{O}})_{mj} \right) \mathbf{x}_j.
\end{equation}
This expansion has a clear topological interpretation: the term $(\mathbf{S}_{\mathcal{U}\mathcal{U}}^k)_{im} (\mathbf{S}_{\mathcal{U}\mathcal{O}})_{mj}$ represents a path of length $k+1$ starting from an observed node $v_j$, passing through a neighbor $v_m \in \mathcal{U}$, and traversing $k$ hops within the missing set $\mathcal{U}$ to reach $v_i$. If $v_i$ resides in a component disjoint from $\mathcal{O}$, there exists no path connecting any observed node $v_j$ to $v_i$, which leads to $\hat{\mathbf{x}}_i = \mathbf{0}$. Assuming the ground truth $\mathbf{x}_i \neq \mathbf{0}$, the imputation error is maximized. Thus, structural reachability is a prerequisite for valid diffusion-based imputation. 
\end{proof}

Lemma \ref{lemma:reachability} indicates that optimizing the diffusion equation is invalid for topologically isolated components. To mitigate this issue, a naive strategy is to randomly link node pairs across different components. However, this approach ignores the semantic consistency of nodes, inevitably introducing noise that damages the quality of imputed features \cite{rossi2022unreasonable}.

To alleviate the unreachability problem while maintaining high reliability, we propose \textbf{Global Structural Augmentation (GSA)}. Instead of random linking, GSA leverages available supervision signals and high-confidence feature similarities to establish semantic bridges. Specifically, as outlined in Algorithm \ref{alg:gsa}, for labeled nodes, we strictly filter global neighbors based on label consistency. For unlabeled nodes, we only retain links with feature similarity exceeding a threshold $\delta$. By constructing the augmented graph $\tilde{\mathcal{G}}$, GSA introduces long-range edges that connect potentially disjoint components. Through this selective linking strategy, GSA provides a structurally enhanced initialization that permits information flow across previously disconnected regions while minimizing the introduction of noisy edges.

\subsection{Stage II: Self-Supervised Manifold Learning}
\label{sec:manifold_learning}

After GSA, we obtain an augmented graph $\tilde{\mathcal{G}}$. While it establishes long-range semantic bridges to alleviate topological isolation, the features initialized by diffusion on $\tilde{\mathcal{G}}$ still suffer from over-smoothing problem. To recover the latent semantic patterns (i.e. the distribution of the real features), we employ a graph-based Masked Autoencoder (MAE) framework to learn the structurally decoupled feature distribution with the following strategy. 

\subsubsection{Dual-Perturbation Strategy}
Standard MAEs typically use binary masking on raw features. However, we argue that this setting is insufficient for missing feature imputation for two reasons. First, the inputs are already smoothed results from propagation, making binary masking a trivial reconstruction task that fails to capture complex dependencies. Second, standard training assumes a static graph structure, leaving the model vulnerable to the structural shifts inherent in inductive tasks. To address these issues, \DART incorporates a dual-perturbation strategy to perceive the distribution of real features that are not strongly correlated with the structure.

\spara{Structural Perturbation for Inductive Robustness} 
In inductive settings, the local topology of a node $v_i$ in the inference graph $\mathbf{A}_{test}$ often differs from that in $\mathbf{A}_{train}$. If the model over-relies on the specific edges in $\mathbf{A}_{tr}$, it will fail to generalize. To force the model to learn representations invariant to local topological changes, we randomly drop edges from the training graph to generate a perturbed view $\mathbf{A}_{drop}$ in each epoch. The input features are then generated by propagating raw features on this partial structure:
\begin{equation}
    \mathbf{X}_{in} = \tilde{\mathbf{A}}_{drop} \mathbf{X}_{in} \odot (\mathbf{1} - \mathbf{M}) + \mathbf{X}_{tr} \odot \mathbf{M},
\end{equation}
where $\tilde{\mathbf{A}}_{drop}$ is the normalized adjacency matrix, and $\mathbf{M}$ is the feature-level binary mask. This equation represents the steady state where information has been fully propagated across the perturbed graph while preserving the observed values. By training on incomplete structures $\mathbf{A}_{drop}$, it forces the encoder to reduce its dependency on specific edges and focus more on the underlying feature distribution.

\spara{Continuous Semantic Perturbation} 
Instead of binary masking, we apply a Gaussian-based continuous masking strategy. Since the FP-imputed inputs $\mathbf{X}_{in}$ are dense continuous values, binary masking (setting to zero) creates an artificial distribution gap. In contrast, adding continuous noise simulates the natural uncertainty of the manifold. We generate the perturbed input $\tilde{\mathbf{X}}$ as:
\begin{equation}
    \mathbf{W} \sim \mathcal{N}(0, 1), \quad \tilde{\mathbf{X}} = \mathbf{X}_{in} \odot \sigma(\mathbf{W} + \delta),
\end{equation}
where $\sigma$ is the sigmoid function and $\delta$ controls the noise level. According to our theoretical analysis in Appendix \ref{sec:theory}, reconstructing representation from continuous noise forces the model to learn the vector field pointing towards the high-density regions of the data distribution. This effectively prevents the model from overfitting of the smoothed values and enables the recovery of high-frequency discriminative details.

\subsubsection{Manifold Learning Objective}
To guide the model learn the correct semantic manifold, we set the features propagated on the GSA graph $\tilde{\mathcal{G}}$ as the reconstruction target, denoted as $\mathbf{X}_{target} = \mathtt{FP}(\mathbf{X}_{tr}, \tilde{\mathcal{G}})$. Let $\hat{\mathbf{X}} = \mathrm{Decoder}(\mathrm{Encoder}(\tilde{\mathbf{X}}))$ be the reconstructed output from the perturbed input. We optimize the model using the Scaled Cosine Error (SCE) loss:
\begin{equation}
    \mathcal{L} = \frac{1}{|\mathcal{V}|}\sum_{v_i \in \mathcal{V}} \left(1 - \frac{\hat{\mathbf{x}}_i^\top \mathbf{x}_{target, i}}{\|\hat{\mathbf{x}}_i\| \cdot \|\mathbf{x}_{target, i}\|}\right)^{\gamma},
\end{equation}
where $\gamma \ge 1$ is a focusing parameter. By mapping the input from a structurally unstable view ($\mathbf{A}_{drop}$) to a structurally augmented target ($\tilde{\mathcal{G}}$), the model is forced to ignore noise in local structure and capture invariant global semantic patterns (i.e., real-feature distribution and structural invariance). 

\subsection{Stage III: Test-Time Distribution Rectification}
\label{sec:inference}

Standard frameworks typically use the trained encoder to generate embeddings directly. However, as defined in Eq. \eqref{eq:delta}, the structural difference in inductive tasks introduces a distribution shift, meaning the FP output on the test graph deviates from the training manifold.

To address this, we propose the test-time rectification stratgy. We interpret the trained autoencoder not just as a feature extractor, but as a manifold projector. The inference proceeds in two steps:

\spara{Step 1: Biased Estimation} We first perform feature propagation on the test graph to obtain an initial estimation. Due to the structural shift, this estimation is biased: $\hat{\mathbf{X}}_{bias} = \mathbf{A}_{test} \mathbf{X}_{in} \odot (\mathbf{1} - \mathbf{M}) + \mathbf{X}_{tr} \odot \mathbf{M}$.

\spara{Step 2: Manifold Projection} We feed $\hat{\mathbf{X}}_{bias}$ into the full encoder-decoder network. Since the model has been trained (via structural perturbation) to reconstruct valid semantics from corrupted structures, it effectively filters out the structural noise in $\hat{\mathbf{X}}_{bias}$:
\begin{equation}
    \mathbf{X}_{rec} = \mathrm{Decoder}(\mathrm{Encoder}(\hat{\mathbf{X}}_{bias})).
\end{equation}
Finally, the rectified features $\mathbf{X}_{rec}$ are passed through the encoder to generate the embeddings $\mathbf{Z} = \mathrm{Encoder}(\mathbf{X}_{rec})$. This re-projection ensures that the input to the downstream classifier lies on the valid feature manifold learned during training, significantly bridging the inductive gap.

\subsection{Theoretical Analysis of the \DART Framework}
\label{sec:theory}

In this section, we provide a rigorous theoretical analysis of \DART, grounding its effectiveness in the \textit{Manifold Hypothesis}. We demonstrate that the two core components of \DART—the self-supervised learner and the test-time rectification—mathematically correspond to score matching and manifold projection, respectively.

\spara{Score Matching for Diversity Recovery}
The over-smoothing issue in GNNs can be viewed as the collapse of the feature distribution $p_{data}(\mathbf{x})$ towards a low-frequency subspace. We show that the Masked Autoencoder (MAE) in \DART counteracts this by learning the score function of the data distribution. Formally, let the corruption process (Gaussian masking) be modeled as $\tilde{\mathbf{x}} = \mathbf{x} + \boldsymbol{\epsilon}$, where $\boldsymbol{\epsilon} \sim \mathcal{N}(\mathbf{0}, \sigma^2 \mathbf{I})$ is the injected noise vector. The MAE training objective minimizes the expected denoising error:
\begin{equation}
    \mathcal{L}_{\text{MAE}} = \mathbb{E}_{\mathbf{x} \sim p_{data}} \mathbb{E}_{\boldsymbol{\epsilon}} \left[ \| \mathbf{x} - r_\theta(\tilde{\mathbf{x}}) \|_2^2 \right],
\end{equation}
where $r_\theta(\cdot)$ denotes the encoder-decoder function parameterized by $\theta$. Following the theory of Denoising Autoencoders (DAE) \cite{alain2014regularized}, the optimal minimizer $r^*(\cdot)$ asymptotically approximates the gradient of the log-density:
\begin{equation}
    \label{eq:score_function}
    r^*(\mathbf{x}) = \mathbf{x} + \sigma^2 \nabla_{\mathbf{x}} \log p_{data}(\mathbf{x}) + o(\sigma^2),
\end{equation}
where $o(\sigma^2)$ represents the higher-order term that approaches zero faster than $\sigma^2$, the gradient term $\nabla_{\mathbf{x}} \log p_{data}(\mathbf{x})$ defines a vector field pointing towards the high-density regions of the manifold \cite{hyvarinen2005estimation}. This derivation implies that \DART learns a vector field pointing towards the high-density modes of the manifold $\mathcal{M}$. While structural propagation acts as a low-pass filter that smooths the semantics, Eq. (\ref{eq:score_function}) shows that the learned MAE model could effectively push representations back towards high-probability modes, thereby recovering semantic diversity.

\spara{Rectification as Manifold Projection}
In inductive settings, the structural change introduces a bias into the FP-imputed features. We model the biased estimation $\hat{\mathbf{x}}_{fp}$ as a perturbation of the ground truth $\mathbf{x}_{gt} \in \mathcal{M}$:
\begin{equation}
    \hat{\mathbf{x}}_{fp} = \mathbf{x}_{gt} + \boldsymbol{\delta}_{\perp},
\end{equation}
where $\boldsymbol{\delta}_{\perp}$ represents the structural noise component orthogonal to the manifold tangent space, which causes the distribution shift. During inference, \DART applies the learned function $r^*(\cdot)$ to this biased input. Substituting $\hat{\mathbf{x}}_{fp}$ into Eq. (\ref{eq:score_function}), the rectification process yields:
\begin{equation}
    \hat{\mathbf{x}}_{rec} = r^*(\hat{\mathbf{x}}_{fp}) \approx \hat{\mathbf{x}}_{fp} + \sigma^2 \nabla_{\mathbf{x}} \log p_{data}(\hat{\mathbf{x}}_{fp}).
\end{equation}
Geometrically, the term $\nabla_{\mathbf{x}} \log p_{data}(\hat{\mathbf{x}}_{fp})$ represents the direction of steepest ascent on the probability landscape. Since $\hat{\mathbf{x}}_{fp}$ lies in a low-density region due to the shift, this gradient term points directly towards the nearest high-density region on $\mathcal{M}$. Consequently, the two-step inference approximates a gradient ascent step that effectively acts as a projection operator:
\begin{equation}
    \hat{\mathbf{x}}_{rec} \approx \hat{\mathbf{x}}_{fp} - \mathcal{P}_{\perp}(\boldsymbol{\delta}_{\perp}) \longrightarrow \mathcal{P}_{\mathcal{M}}(\hat{\mathbf{x}}_{fp}),
\end{equation}
where $\mathcal{P}_{\mathcal{M}}$ denotes the projection onto the intrinsic manifold. Thus, \DART mathematically functions as a manifold projector, filtering out the off-manifold structural noise $\boldsymbol{\delta}_{\perp}$ induced by the inductive feature distribution gap.

\section{Experiments}
\label{sec:experiments}

\begin{table}[t]
\centering
\caption{Dataset statistics (D: Graph Density, C: Clustering coefficient).}
\vspace{-1mm}
\renewcommand{\arraystretch}{0.95}%
\label{tab:datasets}

  \begin{tabular}{lcccc}
  \toprule
    Dataset    & Nodes   & Edges   & D ($10^{-3}$) & C\\
  \midrule
    Cora       & 2,708   & 10,566       &  1.48 & 0.261 \\
    Citeseer   & 3,327   & 9,228       &  0.85 & 0.141 \\
    PubMed     & 19,717  & 88,651       &  0.22 & 0.060 \\
    Ogbn-Arxiv & 169,343 & 1,166,243    &  0.08 & 0.118 \\
    Flickr     & 89,250  & 899,756       &  0.22 & 0.033 \\
    Reddit     & 232,965 & 23,213,838   &  0.42 & 0.240 \\
    Sailing    & 5,014   & 55,458      &  2.21 & 0.323 \\
  \bottomrule
  \end{tabular}
\vspace{-3mm}
\end{table}

In this section, we evaluate \DART compared with eight baselines across two tasks: node classification and link prediction. For node classification, we conduct experiments under both transductive and inductive settings. For link prediction, we evaluate all the methods as specifically demonstrated in Section \ref{subsec:link_prediction}. All experiments are conducted on a single machine with an Intel Xeon 8377C CPU, an NVIDIA RTX 3090 GPU (24GB), and 1TB of RAM.

\begin{table*}[t]
\caption{Transductive node classification results (Sailing with 80.4\% missing rate, others with 90\%).}
\label{tab:transductive}
\centering
\resizebox{\textwidth}{!}{%
  \renewcommand{\arraystretch}{1.05}%
  \begin{small}

  \begin{tabular}{c | c | c | c | c | c | c | c | c | c | c}
  \toprule
  \multirow{1}{*}{Missing Type} & \multirow{1}{*}{Dataset} & \multicolumn{1}{c|}{GAT} & \multicolumn{1}{c|}{GCN} & \multicolumn{1}{c|}{GCNMF} & \multicolumn{1}{c|}{PaGNN} & \multicolumn{1}{c|}{FP} & \multicolumn{1}{c|}{PCFI} & \multicolumn{1}{c|}{ASD-VAE} & \multicolumn{1}{c|}{DRI} & \multicolumn{1}{c}{\DART}\\
  \midrule 
  \multirow{4}{*}{Uniform} 
    & Cora        & 66.06 $\pm$ 1.37 & 62.02 $\pm$ 0.95 & 42.94 $\pm$ 2.36 & 65.30 $\pm$ 1.31 & 74.98 $\pm$ 0.88 & \underline{76.68 $\pm$ 1.55} & 71.94 $\pm$ 0.76 & 72.88 $\pm$ 0.48 & \textbf{79.48 $\pm$ 0.30} \\
    & Citeseer    & 47.68 $\pm$ 1.30 & 44.78 $\pm$ 2.09 & 30.14 $\pm$ 4.68 & 49.36 $\pm$ 1.78 & 58.90 $\pm$ 1.00 & \underline{60.22 $\pm$ 0.80} & 56.94 $\pm$ 0.94 & 57.46 $\pm$ 1.04 & \textbf{62.64 $\pm$ 0.74} \\
    & PubMed      & 67.38 $\pm$ 1.24 & 67.24 $\pm$ 1.84 & 45.00 $\pm$ 3.04 & 69.44 $\pm$ 1.01 & 75.80 $\pm$ 0.62 & \underline{76.74 $\pm$ 0.47} & OOM & 70.88 $\pm$ 0.54 & \textbf{78.88 $\pm$ 0.40} \\
    & Ogbn-Arxiv  & 58.74 $\pm$ 0.21 & 60.76 $\pm$ 0.28 & 57.36 $\pm$ 0.74 & 62.49 $\pm$ 0.20 & 67.84 $\pm$ 0.40 & \underline{68.27 $\pm$ 0.18} & OOM & OOM & \textbf{69.98 $\pm$ 0.22} \\
  \midrule
  \multirow{4}{*}{Structural} 
    & Cora        & 54.88 $\pm$ 1.43 & 53.18 $\pm$ 2.19 & 24.80 $\pm$ 6.72 & 55.26 $\pm$ 2.31 & 73.06 $\pm$ 1.86 & \underline{73.98 $\pm$ 1.61} & 71.00 $\pm$ 1.01 & 72.02 $\pm$ 0.56 & \textbf{77.12 $\pm$ 0.82} \\
    & Citeseer    & 40.52 $\pm$ 1.90 & 37.30 $\pm$ 1.95 & 20.02 $\pm$ 2.53 & 40.34 $\pm$ 2.26 & 55.44 $\pm$ 1.44    & \underline{59.38 $\pm$ 1.02} & 54.52 $\pm$ 1.39 & 56.22 $\pm$ 0.76 & \textbf{60.08 $\pm$ 0.42} \\
    & PubMed      & 63.28 $\pm$ 1.48 & 62.10 $\pm$ 2.00 & 40.46 $\pm$ 4.15 & 64.74 $\pm$ 1.09 & 73.78 $\pm$ 0.35 & \underline{75.08 $\pm$ 2.40} & OOM & 68.96 $\pm$ 0.32 & \textbf{78.02 $\pm$ 0.59} \\
    & Ogbn-Arxiv  & 60.97 $\pm$ 0.33 & 56.83 $\pm$ 0.43 & 34.89 $\pm$ 4.36 & 60.70 $\pm$ 0.42 & \underline{67.77 $\pm$ 0.15} & 67.73 $\pm$ 0.28 & OOM & OOM & \textbf{68.24 $\pm$ 0.29} \\
    \midrule
    \multirow{1}{*}{Natural} 
      & Sailing     &  63.07 $\pm$ 1.47 & 64.45 $\pm$ 0.62 & 48.66 $\pm$ 2.53 & 67.47 $\pm$ 1.11 & 66.91 $\pm$ 0.90 & \underline{68.45 $\pm$ 0.36} & 66.71 $\pm$ 1.02 & 64.62 $\pm$ 0.96 & \textbf{70.88 $\pm$ 0.96} \\
  \bottomrule
  \end{tabular}

  \end{small}
}
\vspace{-2mm}
\end{table*}

\subsection{Datasets and Baselines}

To verify the effectiveness of \DART in real-world scenarios, we propose a new dataset for missing feature imputation called \textbf{Sailing}, which is collected from Danish Nautical Data Records~\cite{dma}. We select a subset from the original data and construct the graph by dividing the grid according to latitude and longitude. Trajectory points in the same grid or adjacent grids are connected as edges. Appendix \ref{sec:datasets-statistics} reports more construction details. After the construction, the Sailing dataset naturally has 80.4\% missing features, providing a new benchmark dataset for missing feature imputation methods to validate their effectiveness in real-world scenarios. Next, we evaluate the transductive node classification task on five datasets: Cora, Citeseer, PubMed \cite{sen2008collective}, Ogbn-Arxiv \cite{hu2020open} and Sailing. To validate our effectiveness on inductive tasks, we evaluate the inductive node classification task on three datasets: Reddit, Flickr \cite{zenggraphsaint} and Sailing. The dataset statistics are shown in Table \ref{tab:datasets}.

\DART is compared with eight baselines, including: (1) two standard graph embedding methods: GCN~\cite{kipf2017semisupervised} and GAT~\cite{velickovic2017graph}; and (2) six methods designed for missing feature scenarios: GCNMF~\cite{taguchi2021graph}, PaGNN~\cite{jiang2020incomplete}, FP~\cite{rossi2022unreasonable}, PCFI~\cite{um2023confidencebased}, ASD-VAE~\cite{jiang2024incomplete} and DRI~\cite{liu2025incomplete}. Following previous work \cite{rossi2022unreasonable,um2023confidencebased}, for public datasets, we generate the missing features with \textit{Uniform} and \textit{Structural} missing types as input for all methods. Specifically, uniform missing means that each feature dimension of each node is missing randomly, while structural missing represents that either we observe all features for a node, or we observe none. For the Sailing dataset, we utilize its naturally missing features. For FP and PCFI, we use the same type of GNN as the encoder in \DART for a fair comparison.

\subsection{Experimental Setup}
For the transductive and inductive node classification task, we report the mean and standard error of the test accuracy with running five times. For the link prediction task, AUC and AP scores~\cite{fawcett2006introduction} are used to validate the effectiveness of all the methods following prior work~\cite{song2021dynamic,yang2022scalable,yang2020homogeneous}. All results are under 90\% missing feature rate (except for Sailing, which naturally has an 80.4\% missing feature rate). For \DART, we first simply set the $\tau = 1e9$ (uniform sampling) for GSA and follow previous work \cite{hou2022graphmae} to set the parameters for MAE to avoid excessive parameter tuning. Then we adjust GSA parameters $k$ and $\delta$ to achieve the best result on the validation set, with all details provided in Section \ref{sec:para_sensitivity}. For other baselines, we use their official code to reproduce the results. Note that distinct from prior studies \cite{he2022masked,um2023confidencebased,jiang2024incomplete} that often limit evaluation to the Largest Connected Component (LCC), we use the general GNN evaluation protocol that all experiments are conducted on the \textbf{entire} test set rather than a subset.

\subsection{Node Classification}
\label{subsec:node_classification}

The node classification task aims at predicting unknown labels of nodes in a graph. For the transductive setting, the whole graph structure is available and we need to predict the labels of the test set with training labels. Here we use the official split for Cora, Citeseer, PubMed and Ogbn-Arxiv \cite{kipf2017semisupervised}, and split the Sailing with 40\%-30\%-30\% for train, validation and test. For the inductive setting, only nodes that appear in the training set are available during training, and the whole graph structure is available for inference. We split the Flickr and Reddit with 20\%-40\%-40\% for train, validation and test, and use the same split as transductive setting for Sailing.

\spara{Transductive Results} Table \ref{tab:transductive} reports the results of all the methods on the transductive node classification task with uniform and structural missing types. \DART outperforms all the baselines across every dataset. Specifically, on the four public datasets with uniformly missing features, \DART achieves consistent improvements, surpassing the strongest baseline PCFI by margins ranging from $1.71\%$ to $2.80\%$. Notably, on the real-world Sailing dataset, our method achieves a significant improvement of $2.43\%$ compared to the runner-up PCFI, demonstrating its robustness in handling natural feature sparsity. In scenarios with structurally missing features, \DART maintains its superiority. On the Cora dataset, for instance, it surpasses PCFI by an accuracy margin of $3.14\%$.
Overall, methods utilizing feature imputation generally outperform standard GNNs like GAT and GCN, which suffer performance drops as they assume complete features. An exception is GCNMF, whose performance significantly declines under high missing rates. FP performs relatively well in transductive settings due to its diffusion mechanism, but \DART achieves the best results by effectively integrating feature rectification and structural augmentation.

\spara{Inductive Results} We further evaluate the performance under the inductive setting, as shown in Table \ref{tab:inductive}. Since ASD-VAE is not designed for inductive tasks and runs out-of-memory on large graphs like Reddit, it is excluded from this comparison. The results indicate that baselines relying solely on simple propagation, such as FP and PCFI, often result in suboptimal performance due to the feature distribution shift between training and inference graphs. Similarly, standard GNNs struggle to generalize to unseen nodes with missing features. In contrast, \DART effectively bridges this gap by utilizing masked autoencoders to learn the underlying feature distribution. Consequently, \DART achieves substantial gains across all inductive benchmarks. For example, on the Sailing dataset, \DART outperforms the second-best method PaGNN by a remarkable margin of $12.71\%$ and PCFI by $14.04\%$, validating the effectiveness of our proposed distribution rectification mechanism.

\spara{Robustness to Missing Rates} Additionally, we evaluate the performance across varying missing rates (ranging from 10\% to 90\%) on four datasets: Cora, Citeseer, PubMed, and Ogbn-Arxiv. ASD-VAE and DRI are excluded from large datasets due to out-of-memory issues, and GCNMF is omitted due to its consistently inferior performance. The results, visualized in Figure \ref{fig:results_different_rate}, reveal that most methods not based on feature propagation experience significant performance degradation as the missing rate increases. In contrast, \DART consistently achieves superior performance across different missing rates, underscoring its robustness in handling severe feature sparsity for real-world datasets.

\begin{figure*}[t]
\centering
\captionsetup[subfloat]{captionskip=0.5mm}
\begin{small}
\begin{tabular}{cccc}
\multicolumn{4}{c}{\hspace{-4mm} \includegraphics[width=0.5\linewidth]{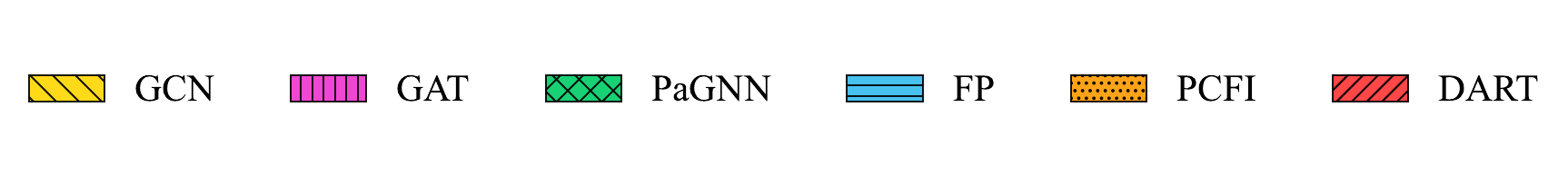}}\vspace{-5mm}  \\
\hspace{-2mm}\subfloat[{\em Cora}]{\includegraphics[width=0.243\linewidth]{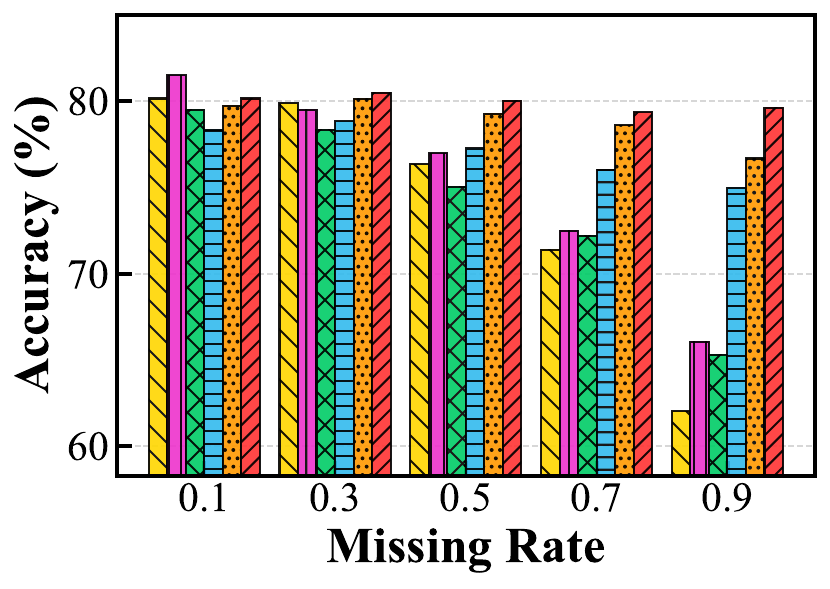}\label{fig:cora}} &
\hspace{-2mm}\subfloat[{\em Citeseer}]{\includegraphics[width=0.243\linewidth]{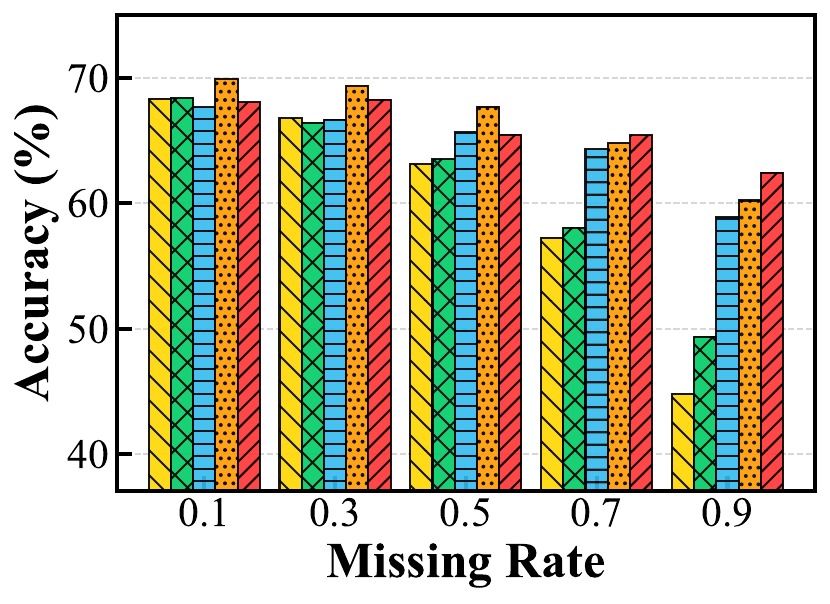}\label{fig:citeseer}} &
\hspace{-2mm}\subfloat[{\em PubMed}]{\includegraphics[width=0.243\linewidth]{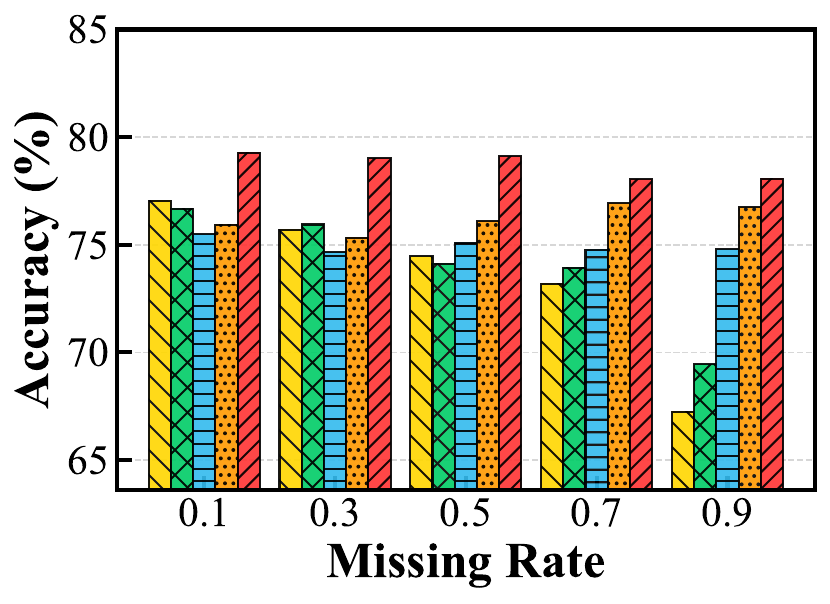}\label{fig:pubmed}} &
\hspace{-2mm}\subfloat[{\em Ogbn-Arxiv }]{\includegraphics[width=0.243\linewidth]{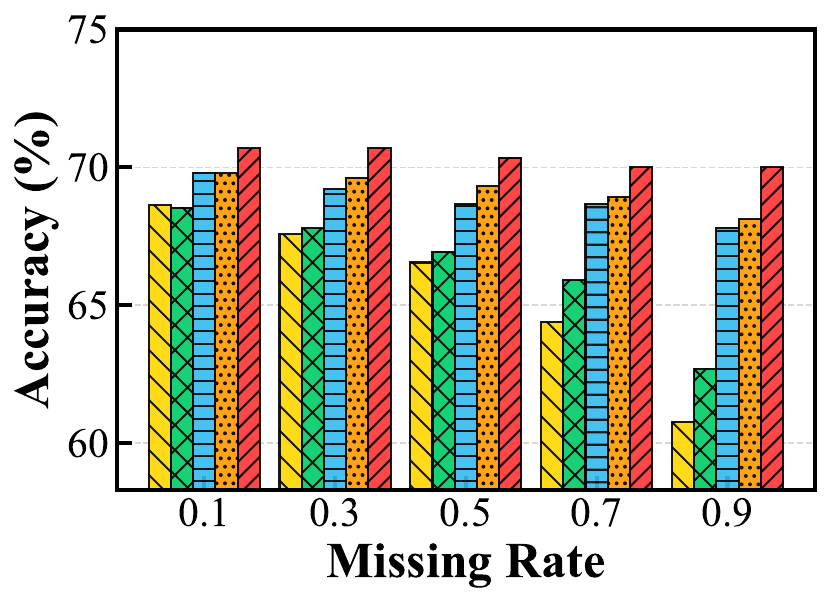}\label{fig:arxiv}}\\
\end{tabular}
\caption{Performance under different missing rates for transductive learning (best viewed in color).}
\vspace{-1mm}
\label{fig:results_different_rate}
\end{small}
\end{figure*}

\begin{table*}[t]
\centering
\caption{Inductive node classification results.}
\vspace{-1mm}

\label{tab:inductive}
  \renewcommand{\arraystretch}{1}%
  \begin{small}
  \begin{tabular}{c | c | c | c | c | c | c | c | c | c}
  \toprule
  \multirow{1}{*}{Missing Type} & \multirow{1}{*}{\makebox[0.08\textwidth]{Dataset}} & \multicolumn{1}{c|}{GAT} & \multicolumn{1}{c|}{GCN} & \multicolumn{1}{c|}{GCNMF} & \multicolumn{1}{c|}{PaGNN} & \multicolumn{1}{c|}{FP} & \multicolumn{1}{c|}{PCFI} & \multicolumn{1}{c|}{DRI} & \multicolumn{1}{c}{\DART} \\
  \midrule
  \multirow{2}{*}{Uniform}
    & Flickr  & 43.90 $\pm$ 0.94 & 44.99 $\pm$ 0.33 & 47.73 $\pm$ 2.71 & \underline{49.46 $\pm$ 0.12} & 48.10 $\pm$ 0.77 & 47.72 $\pm$ 0.47 & 45.14 $\pm$ 0.28 & \textbf{51.98 $\pm$ 0.08} \\
    & Reddit  & 49.90 $\pm$ 2.50 & 84.08 $\pm$ 0.85 & OOM             & \underline{86.77 $\pm$ 0.50} & 86.28 $\pm$ 2.69 & OOM    & OOM    & \textbf{93.88 $\pm$ 0.21} \\
  \midrule
  \multirow{2}{*}{Structural}
    & Flickr  & 42.48 $\pm$ 0.09 & 42.78 $\pm$ 0.07 & 43.09 $\pm$ 1.01 & 45.72 $\pm$ 0.29 & \underline{45.94 $\pm$ 1.59} & 45.25 $\pm$ 0.68 & 43.24 $\pm$ 0.37 & \textbf{50.64 $\pm$ 0.20} \\
    & Reddit  & 77.96 $\pm$ 2.68 & 84.86 $\pm$ 0.97 & OOM             & \underline{90.57 $\pm$ 0.32} & 89.47 $\pm$ 1.91 & OOM   & OOM      & \textbf{92.32 $\pm$ 0.38} \\
    \midrule
    \multirow{1}{*}{Natural}
        & Sailing & 59.64 $\pm$ 0.97 & 61.62 $\pm$ 0.60 & 45.10 $\pm$ 4.84 & \underline{64.85 $\pm$ 0.56} & 63.44 $\pm$ 0.98 & 63.52 $\pm$ 0.21 & 61.86 $\pm$ 0.41 & \textbf{77.56 $\pm$ 0.24} \\
  \bottomrule
  \end{tabular}

  \end{small}
\vspace{-1mm}
\end{table*}

\subsection{Link Prediction}
\label{subsec:link_prediction}
\begin{table*}[t]
\centering
\caption{Link prediction results.}
\vspace{-1mm}
\label{tab:link_pred}
\resizebox{\textwidth}{!}{%
  \renewcommand{\arraystretch}{1.05}%
  \begin{small}
  \begin{tabular}{c | c | c | c | c | c | c | c | c | c | c | c}
  \toprule
  \multirow{2}{*}{} & \multirow{2}{*}{Dataset} & \multicolumn{2}{c|}{GCNMF} & \multicolumn{2}{c|}{GCN} & \multicolumn{2}{c|}{FP} & \multicolumn{2}{c|}{PCFI} & \multicolumn{2}{c}{\DART}\\
  \cmidrule{3-12}
   & & \multicolumn{1}{c|}{AUC} & \multicolumn{1}{c|}{AP} & \multicolumn{1}{c|}{AUC} & \multicolumn{1}{c|}{AP} & \multicolumn{1}{c|}{AUC} & \multicolumn{1}{c|}{AP} & \multicolumn{1}{c|}{AUC} & \multicolumn{1}{c|}{AP} & \multicolumn{1}{c|}{AUC} & \multicolumn{1}{c}{AP} \\
  \midrule 
  \multirow{5}{*}{\rotatebox{90}{\scriptsize{Uniform}}}
    & Cora          & 61.09 $\pm$ 1.54 & 61.73 $\pm$ 2.23 & 71.13 $\pm$ 2.19 & 72.86 $\pm$ 2.62 & \underline{86.65 $\pm$ 0.68} & \underline{89.21 $\pm$ 0.35} & 86.16 $\pm$ 0.89 & 88.91 $\pm$ 0.46 & \textbf{94.62 $\pm$ 0.90} & \textbf{94.28 $\pm$ 0.79} \\
    & Citeseer      & 59.00 $\pm$ 1.72 & 58.54 $\pm$ 2.09 & 72.31 $\pm$ 2.44 & 75.65 $\pm$ 2.27 & \underline{81.18 $\pm$ 1.17} & \underline{86.02 $\pm$ 1.30} & 80.68 $\pm$ 1.38 & 85.44 $\pm$ 1.01 & \textbf{93.90 $\pm$ 0.96} & \textbf{93.02 $\pm$ 0.87} \\
    & PubMed        & 61.62 $\pm$ 0.51 & 69.46 $\pm$ 0.67 & \underline{82.18 $\pm$ 0.77} & 82.33 $\pm$ 0.45 & 81.15 $\pm$ 0.19 & 85.61 $\pm$ 0.19 & 80.72 $\pm$ 1.02 & \underline{85.62 $\pm$ 0.56} & \textbf{97.28 $\pm$ 0.18} & \textbf{97.04 $\pm$ 0.17} \\
    & Ogbn-Arxiv    & 84.87 $\pm$ 1.02 & 87.09 $\pm$ 0.93 & 90.97 $\pm$ 0.43 & 91.52 $\pm$ 0.44 & 94.71 $\pm$ 0.31 & 95.20 $\pm$ 0.28 & \underline{95.72 $\pm$ 0.10} & \underline{96.04 $\pm$ 0.10} & \textbf{98.78 $\pm$ 0.08} & \textbf{98.88 $\pm$ 0.07} \\
  \midrule
  \multirow{4}{*}{\rotatebox{90}{\scriptsize{Structural}}}
    & Cora          & 58.52 $\pm$ 1.19 & 63.83 $\pm$ 1.49 & 67.86 $\pm$ 1.65 & 70.15 $\pm$ 1.55 & \underline{85.24 $\pm$ 1.55} & 87.54 $\pm$ 1.26 & 85.11 $\pm$ 1.27 & \underline{87.96 $\pm$ 1.02} & \textbf{94.56 $\pm$ 1.20} & \textbf{93.92 $\pm$ 1.61} \\
    & Citeseer      & 56.39 $\pm$ 1.94 & 63.32 $\pm$ 1.75 & 62.93 $\pm$ 2.25 & 66.21 $\pm$ 1.46 & 80.07 $\pm$ 1.31 & 84.06 $\pm$ 1.42 & \underline{81.53 $\pm$ 1.28} & \underline{85.00 $\pm$ 1.44} & \textbf{93.78 $\pm$ 0.48} & \textbf{93.04 $\pm$ 0.82} \\
    & PubMed        & 63.25 $\pm$ 0.37 & 72.54 $\pm$ 0.30 & \underline{86.30 $\pm$ 0.55} & 84.81 $\pm$ 0.29 & 79.59 $\pm$ 0.75 & 84.81 $\pm$ 0.47 & 82.93 $\pm$ 0.41 & \underline{86.62 $\pm$ 0.37} & \textbf{97.02 $\pm$ 0.19} & \textbf{96.70 $\pm$ 0.21} \\
    & Ogbn-Arxiv    & 84.97 $\pm$ 1.02 & 87.26 $\pm$ 0.87 & 90.97 $\pm$ 0.43 & 91.53 $\pm$ 0.43 & 94.69 $\pm$ 0.29 & 95.17 $\pm$ 0.25 & \underline{95.71 $\pm$ 0.09} & \underline{96.03 $\pm$ 0.10} & \textbf{98.66 $\pm$ 0.11} & \textbf{98.68 $\pm$ 0.08} \\
    \midrule
    & Sailing       & 95.78 $\pm$ 0.20 & 95.25 $\pm$ 0.24 & 96.03 $\pm$ 0.20 & 95.21 $\pm$ 0.32 & 97.89 $\pm$ 0.12 & 97.67 $\pm$ 0.17 & \underline{97.97 $\pm$ 0.08} & \underline{97.80 $\pm$ 0.09} & \textbf{98.52 $\pm$ 0.12} & \textbf{98.38 $\pm$ 0.14} \\
  \bottomrule
  \end{tabular}
  \end{small}
}
\vspace{-1mm}
\end{table*}

The link prediction task involves predicting missing or unknown edges within a graph. We conduct experiments on five datasets: Cora, Citeseer, PubMed, Ogbn-Arxiv, and Sailing. For Cora, Citeseer, PubMed and Ogbn-Arxiv, we randomly sample positive and negative edges for validation (5\%) and testing (10\%). For the Sailing dataset, we adopt a split of 60\%/10\%/30\% for training, validation, and test. Following previous work \cite{um2023confidencebased}, we sample the same number of negative edges as positive edges to train a Graph Autoencoder (GAE) \cite{kipf2016variational} for all methods to predict the link. Performance is measured using the Area Under the Curve (AUC) and Average Precision (AP) metrics.

\spara{Results} Table \ref{tab:link_pred} summarizes the link prediction performance under uniform and structural missing settings, as well as the naturally missing setting for Sailing. \DART consistently achieves superior performance across all datasets and metrics. Specifically, in the uniform missing setting, \DART demonstrates substantial improvements over the strongest baselines. For instance, on the PubMed dataset, it outperforms the runner-up by a significant margin of $15.10\%$ in terms of AUC. Across the four public datasets, the AUC improvements range from $3.06\%$ to $15.10\%$, validating the model's ability to reconstruct semantically meaningful features that align well with the graph topology. In the structural missing setting, \DART maintains its dominance. Notably, on the Citeseer dataset, it achieves an AUC of $93.78\%$, surpassing the best baseline by $12.25\%$. This indicates that even when node features are entirely absent, \DART can effectively leverage the structural information to recover latent semantics.
On the real-world Sailing dataset, which features naturally missing data, \DART also achieves the highest performance with an AUC of $98.52\%$ and AP of $98.38\%$, slightly outperforming the competitive PCFI baseline. This confirms the robustness and practical value of \DART in handling complex, real-world missing patterns.

\subsection{Ablation Studies}
\label{sec:ablation}
To further investigate the contribution of each mechanism in \DART, we conduct ablation studies on three representative datasets: Cora and PubMed (Transductive), and Flickr (Inductive). We compare the full model against three variants: (1) \textbf{w/o Rec}: removing the rectification module; (2) \textbf{w/o GSA}: removing the Global Structural Augmentation strategy; and (3) \textbf{w/o Gaussian Mask}: replacing the Gaussian noise masking with no corruption. The results are summarized in Table \ref{tab:ablation_node_rec_cll}.

\begin{table}[t]
\centering
\caption{Ablation studies of each mechanism of \DART on different datasets.}
\label{tab:ablation_node_rec_cll}
\resizebox{\linewidth}{!}{%
\begin{tabular}{l | c | c | c }
    \toprule
    \multicolumn{1}{c}{Variant} & \multicolumn{1}{c}{Cora}  & \multicolumn{1}{c}{PubMed} & \multicolumn{1}{c}{Flickr}  \\
    \midrule
    w/o Rec              & 78.88 $\pm$ 0.86  & 77.96 $\pm$ 0.48 & 50.48 $\pm$ 0.28   \\
    w/o GSA              & 77.10 $\pm$ 1.02  & 77.98 $\pm$ 0.59 & 51.93 $\pm$ 0.11   \\
    w/o Gaussian Mask    & 78.16 $\pm$ 1.03  & 77.92 $\pm$ 0.43 & 51.64 $\pm$ 0.14 \\
    \textbf{\DART (Full)} & \textbf{79.48 $\pm$ 0.30}  & \textbf{78.88 $\pm$ 0.40} & \textbf{51.98 $\pm$ 0.08}  \\
    \bottomrule
\end{tabular}
}
\vspace{-3mm}
\end{table}

\spara{Effect of Rectification} 
The rectification module is designed to rectify the feature distribution shift. As shown in Table \ref{tab:ablation_node_rec_cll}, removing this component leads to a consistent performance drop. In the inductive setting (Flickr), the accuracy drops by $1.50\%$, verifying the necessity of rectification in aligning the propagated features with the true data manifold. Similarly, in the transductive setting, we observe improvements ranging from $0.60\%$ to $0.92\%$, attributing to the denoising effect that refines the feature representations.

\spara{Effect of Global Structural Augmentation (GSA)} 
GSA aims to mitigate topological isolation. The results indicate that GSA is particularly critical for sparse graphs like Cora, where removing it causes a significant performance decline of $2.38\%$. On PubMed, the drop is moderate ($0.90\%$). In contrast, on the dense Flickr graph, the impact is negligible ($0.05\%$). This aligns with our analysis that dense graphs naturally possess high reachability, whereas GSA serves as a vital bridge for information flow in sparse scenarios, ensuring the robustness of \DART across varying graph densities.

\spara{Effect of Gaussian Masking} 
We introduce Gaussian masking to force the autoencoder to learn the gradient field of the data distribution (i.e., the score function). The ablation results show that removing Gaussian masking (w/o Gaussian Mask) leads to performance degradation across all datasets, with drops of $1.32\%$ on Cora and $0.34\%$ on Flickr. This confirms that simple rectification without proper noise injection may lead to learning trivial identity mappings. The Gaussian noise perturbation is essential for learning a robust manifold projector that can effectively pull off-manifold features back to high-density regions.

\subsection{Parameter Sensitivity Analysis}
\label{sec:para_sensitivity}
We investigate the sensitivity of \DART to two key hyperparameters in the GSA module: size $K$ and the similarity threshold $\delta$. We conduct experiments on the Cora dataset with $K \in \{5, 10, 20, 40\}$ and $\delta \in \{0.4, 0.6, 0.8, 1.0\}$.

The results indicate that \DART is sensitive to the size $K$. Specifically, it achieves the highest accuracy when $K=10$ and $\delta=0.6$. Performance degrades when $K$ is too small (e.g., $K=5$) as it limits the semantic receptive field, while an excessively large $K$ (e.g., $K=40$) introduces noise by connecting to semantically irrelevant nodes. Regarding the threshold $\delta$, moderate values generally yield superior robustness by effectively filtering out low-confidence links. Therefore, a moderate $K$ around $10$ and $\delta$ larger than $0.6$ is recommended to balance information gain and noise control.

\begin{figure}[t] 
\centering 
\includegraphics[width=0.8\linewidth]{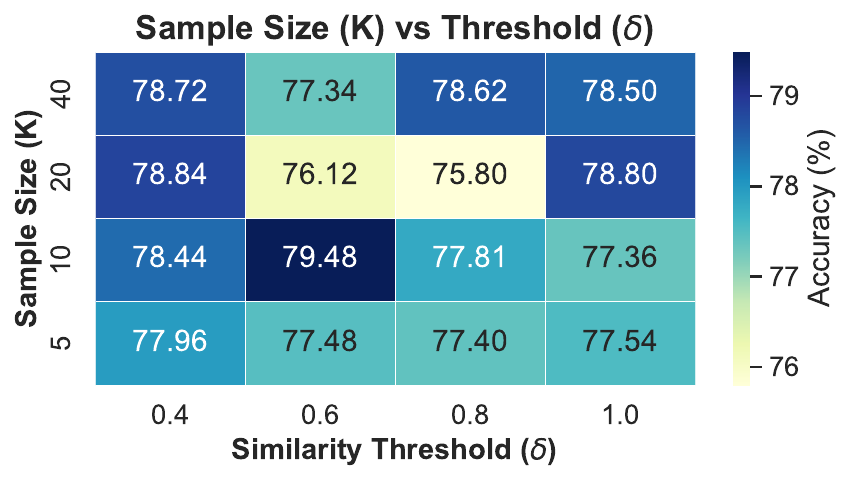}  
\vspace{-3mm}
\caption{Sensitive analysis of $k$ and $\delta$.} \label{fig:sensivity}
\vspace{-3mm}
\end{figure}

\begin{figure*}[t]
    \centering
    \captionsetup[subfloat]{captionskip=0.5mm}
    \begin{small}
    \begin{tabular}{cccc}
    % \hspace{-4mm}\subfloat[{\em Wiki}]{\includegraphics[width=0.26\linewidth]{./figure/class/wiki-eps-converted-to.pdf}\label{fig:acc-class-wiki}} &
    \hspace{-2mm}\subfloat[Raw features (Cora)]{\includegraphics[width=0.23\linewidth]{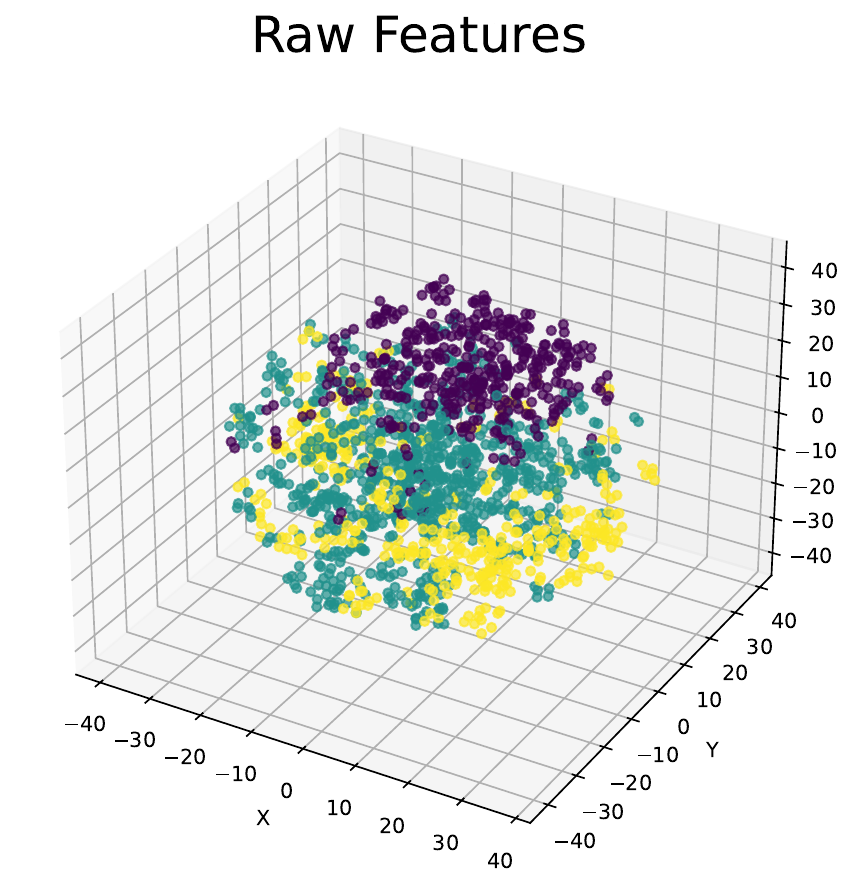}\label{fig:truecora}} &
    \hspace{-2mm}\subfloat[Rectification (Cora)]{\includegraphics[width=0.23\linewidth]{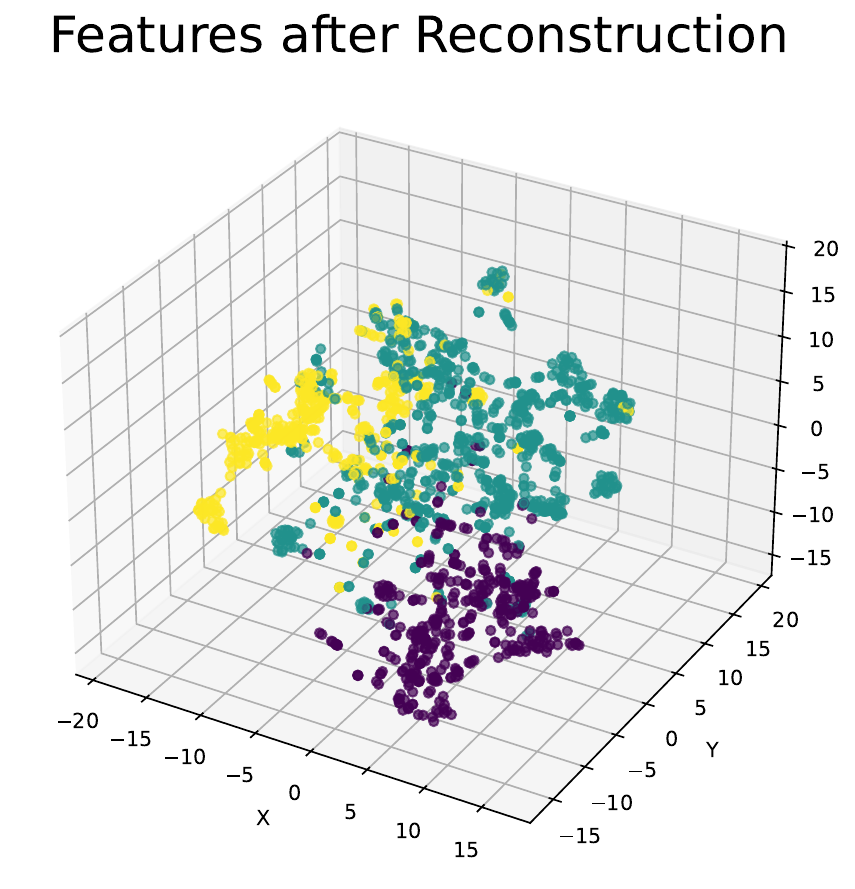}\label{fig:reccora}} &
    \hspace{-2mm}\subfloat[Raw features (Flickr)]{\includegraphics[width=0.23\linewidth]{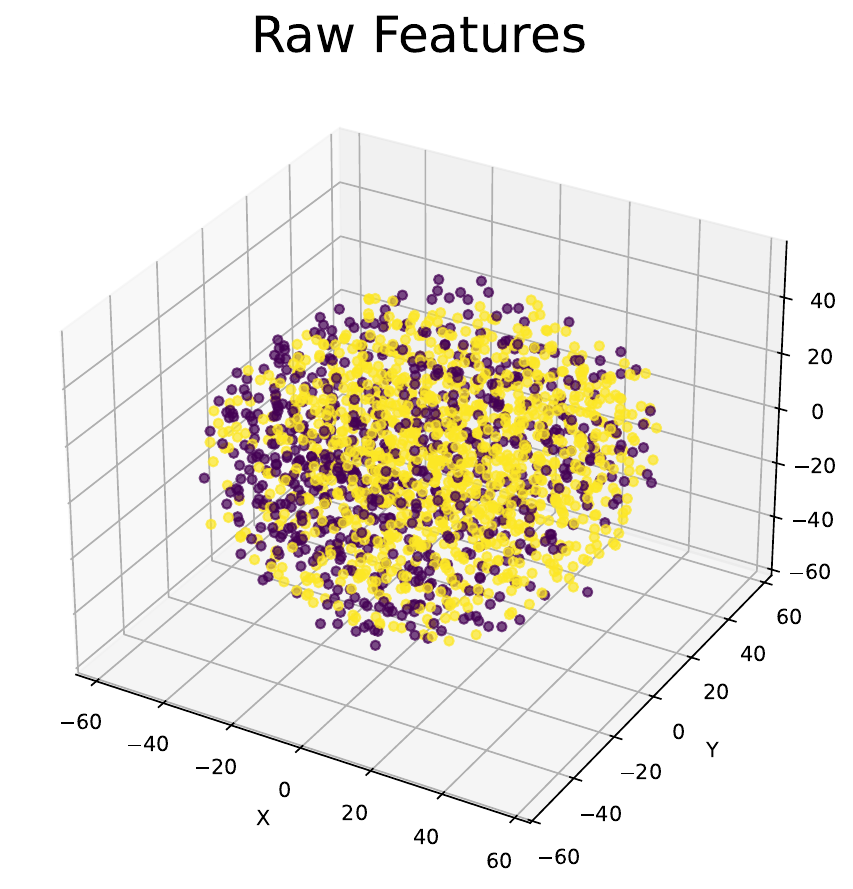}\label{fig:trueflickr}} &
    \hspace{-2mm}\subfloat[Rectification (Flickr)]{\includegraphics[width=0.23\linewidth]{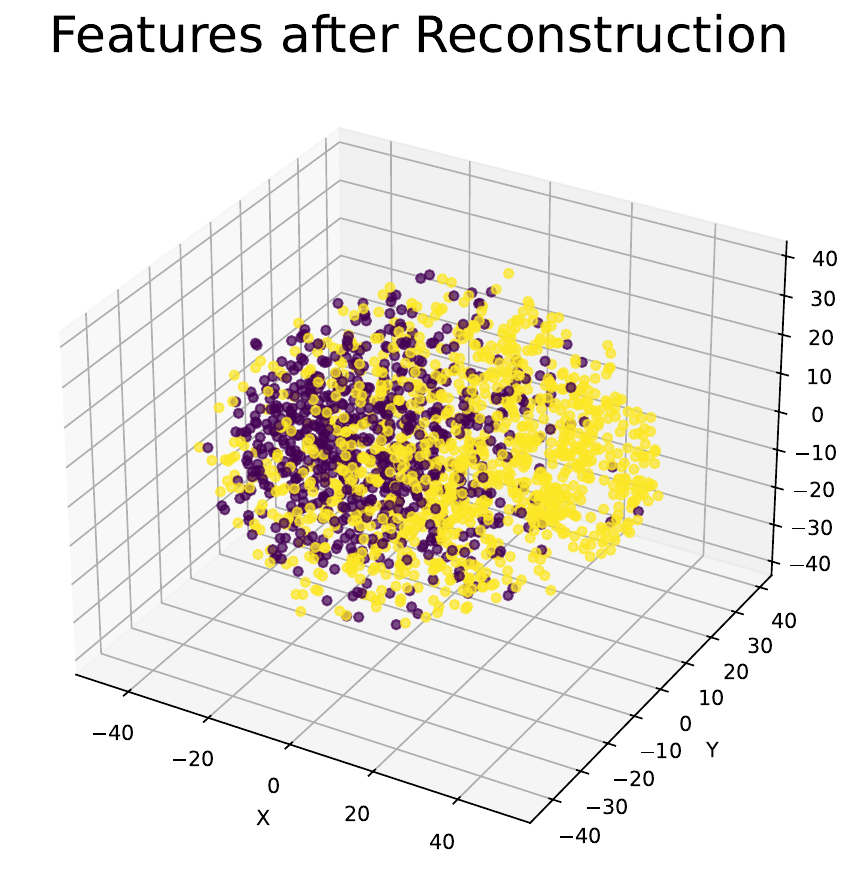}\label{fig:finalflickr}}\\
    \end{tabular}
    \vspace{-2mm}
    \caption{Visualization for transductive (Cora) and inductive (Flickr): raw features and features after rectification.}
    \label{fig:visualization}
    \end{small}
\vspace{-2mm}
\end{figure*}

\subsection{Visualization}
\label{subsec:visulization}

To intuitively verify the effectiveness of our proposed feature rectification mechanism, we employ t-SNE~\cite{van2008visualizing} to project the high-dimensional feature vectors into a low-dimensional space for visualization. Figure \ref{fig:visualization} presents a comparative analysis between the raw features and the rectified features (processed by DART). In the transductive setting on the Cora dataset, the raw features exhibit blurred boundaries with loosely scattered nodes, while the feature distribution becomes significantly more discriminative after rectification. This suggests that DART effectively filters out the noise introduced by topological diffusion, providing clearer decision boundaries for downstream classifiers. Furthermore, we visualize the Flickr dataset to assess performance under the more challenging inductive setting. Due to the large scale of Flickr, we randomly sample 3\% nodes from the two most representative classes. While the raw features suffer from severe overlapping due to distribution shift, the visualization reveals that DART successfully mitigates this issue by projecting these drifting features back onto a structured manifold, resulting in a clear separation between the two classes. This visual evidence strongly supports our claim that the MAE module learns a robust, structure-invariant feature distribution, which is crucial for generalizing to unseen graphs.

\section{Related Work}
\label{sec:relatedwork}

\spara{Missing node features imputation} Some existing studies~\cite{huo2023t2,rossi2022unreasonable,taguchi2021graph,jiang2020incomplete,jiang2024incomplete,you2020handling} explore methods to improve the performance of Graph Neural Networks (GNNs) in handling missing data. SAT~\cite{chen2020learning} introduces a Transformer-like model for feature reconstruction alongside a GNN model to address downstream tasks. GCNMF~\cite{taguchi2021graph} represents missing data using a Gaussian Mixture Model; however, its performance significantly declines when the proportion of missing features is high. PaGNN~\cite{jiang2020incomplete} proposes two novel partial aggregation (reconstruction) functions specifically designed for incomplete graphs. Feature Propagation (FP)~\cite{rossi2022unreasonable}, a general approach for managing missing node features in graph machine learning tasks, includes an initial diffusion-based feature reconstruction step followed by a downstream GNN. ADS-VAE \cite{jiang2024incomplete} applys VAE model to learn the feature imputation, but it needs excessive space consumption and achieves poor performance on complete datasets with multiple connected components. While these methods have shown promising results in transductive graph learning tasks, they are not well-suited for inductive tasks. Also, they are tailored for fully-connected graphs, which makes their performance seriously decline on graphs with many connected components. 

\spara{Inductive Learning on graphs} Transductive GNNs are limited to making predictions on nodes and edges seen during training. In contrast, inductive GNNs can generalize to previously unseen nodes and edges, making them more versatile for real-world applications. GraphSAGE~\cite{hamilton2017inductive} generates embeddings by sampling and aggregating features from a node's local neighborhood, enabling the model to infer embeddings for new nodes that were not present during training. GAT~\cite{velickovic2017graph} assigns different weights to the neighboring nodes during the aggregation process, allowing the model to focus on the most relevant parts of the graph. Some recent studies have developed inductive GNN models for recommender systems~\cite{Zhang2020Inductive,zhang2019star}. \cite{Zeng2020GraphSAINT} propose a graph sampling approach to build subgraphs for training GNNs on large graph datasets.

\spara{Graph Autoencoders} Autoencoders~\cite{hinton1993autoencoders} are designed to reconstruct specific inputs based on the given contexts and do not impose any decoding order, unlike autoregressive methods. Graph Autoencoders mostly adopt structural and feature reconstruction as their objectives~\cite{park2019symmetric, wang2017mgae}. Masked Autoencoders (MAEs)~\cite{he2022masked} are a type of neural network architecture primarily designed for self-supervised learning tasks. GraphMAE~\cite{hou2022graphmae} leverages the inherent structure of graph data to improve learning and representation by masking and predicting parts of the graph during training. 
\section{Conclusion}
\label{sec:conclusion}
In this paper, we introduced \DART, a novel framework that synergizes feature propagation with a distribution-aware rectification mechanism to address missing feature imputation in both transductive and inductive settings. To overcome topological limitations, a Global Structural Augmentation (GSA) strategy is introduced to bridge disjoint components and enhance diffusion reachability. Furthermore, a manifold learner coupled with a test-time rectification scheme is developed to recover semantic diversity and mitigate the feature distribution shift caused by structural changes. To benchmark performance in realistic scenarios, we propose a new maritime dataset Sailing characterized by naturally missing attributes. Extensive experiments on Sailing and six public datasets demonstrate that \DART significantly outperforms state-of-the-art methods across node classification and link prediction tasks, highlighting its robustness and generalizability. In the future work, we expect to expand DART to more extensive scenarios (e.g., heterophilic graphs).

% \spara{Limitations and future directions} Although \DART shows promising results in different datasets, it still has some areas for potential improvement. For example, investigating the extension on heterogeneous graphs or graphs with both structure and features missing could benefit the application of \DART for real-world data. It is also worth exploring whether \DART could apply to other domains such as protein molecule prediction.
\appendix
\section{Appendix}
\label{sec:appendix}

\subsection{Feature distribution shift validation}

According to Section \ref{sec:preliminaries}, the core reason behind the feature distribution shift issue is that the distribution of feature imputed by feature propagation strictly rely on the graph structure, make the difference between training and inference. To validate this, we set the function $\mathcal{F}$ as the Kullback-Leibler Divergence and record the $\Delta_{\mathcal{S}}$ in Table \ref{tab:KL}. It could be seen that \DART can effectively reduce the feature distribution shift.
\begin{table}[h]
\caption{$\Delta_{\mathcal{S}}$ on Flickr, Reddit and Sailing.}
\vspace{-1mm}
\centering
\renewcommand{\arraystretch}{1}
\begin{small}
\begin{tabular}{ c | c | c | c}
\toprule
 \multirow{1}{*}{Dataset} & \multicolumn{1}{c|}{Flickr} & \multicolumn{1}{c}{Reddit} & \multicolumn{1}{c}{Sailing}\\
\midrule 
  FP & 1.08 & 0.16 & 1.96\\
  \DART & \textbf{0.82} ($\downarrow 24.07\%$) & \textbf{0.10} ($\downarrow 37.50\%$) & \textbf{1.25} ($\downarrow 36.22\%$)\\
\bottomrule
\end{tabular}
\end{small}
\label{tab:KL}
\vspace{-3mm}
\end{table} 

\subsection{Construction of Sailing}
\label{sec:datasets-statistics}

\begin{figure}[h]
\centering 
\includegraphics[width=0.9\linewidth]{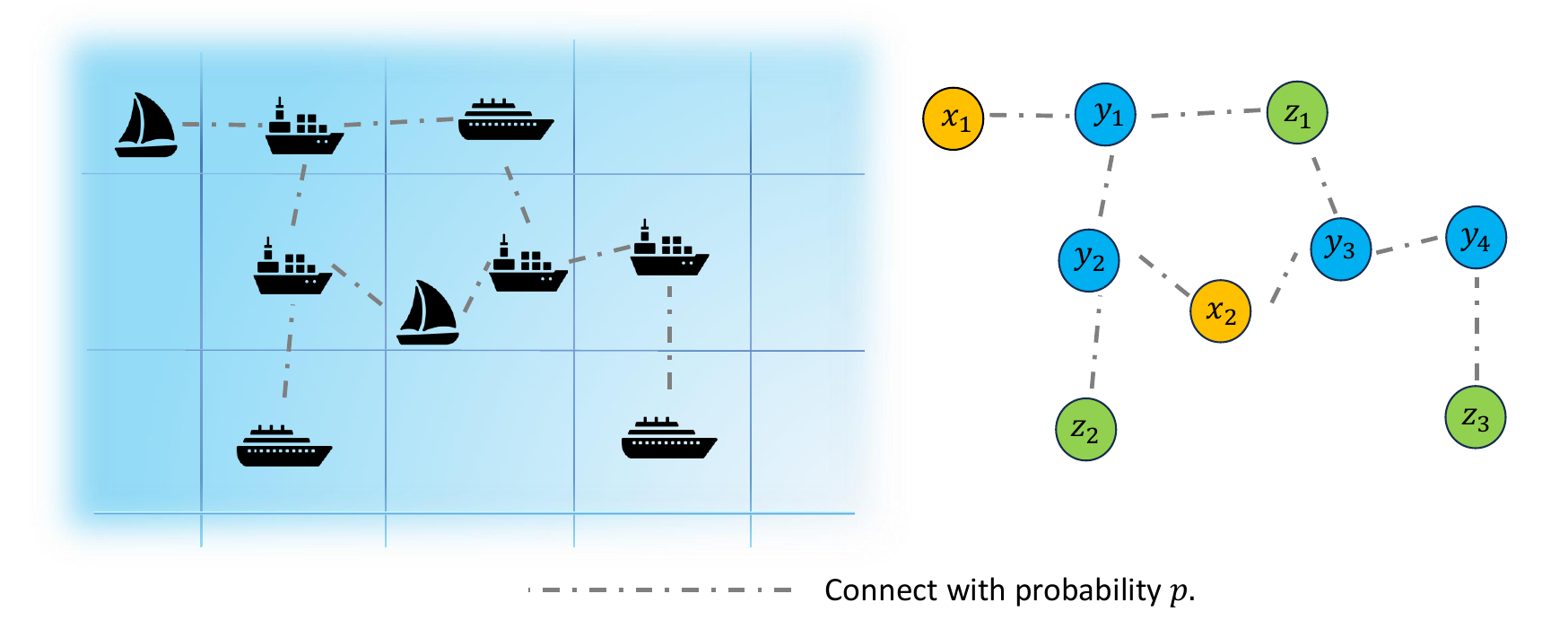}  
\vspace{-1mm}
\caption{The construction of Sailing dataset.} \label{fig:dataset_construction} 
\vspace{-2mm}
\end{figure}

Our original Sailing dataset contains information about multiple ships, with each ship represented as a node in a graph. The navigation status of each ship serves as a label, with a total of n labels. Edges between nodes are constructed based on the ships' longitude and latitude coordinates. We divide the entire ocean into several small areas using these coordinates. If two nodes are in the same area or in adjacent areas, we create an undirected edge between them with a 50\% probability. The process is shown in Figure \ref{fig:dataset_construction}. The dataset includes both qualitative and quantitative features for each ship. Qualitative features are converted into tensor data using one-hot encoding, while quantitative features are normalized. The Sailing dataset naturally contains missing or invalid data, which we mark as 'NaN'.

\end{sloppy}
%%
%% The next two lines define the bibliography style to be used, and
%% the bibliography file.
\bibliographystyle{ACM-Reference-Format}
\bibliography{reference}

%%
%% If your work has an appendix, this is the place to put it.

\end{document}